\documentclass[conference]{IEEEtran}
\IEEEoverridecommandlockouts
\usepackage{cite}
\usepackage{amsmath,amssymb,amsfonts}
\usepackage{algorithmic}
\usepackage{graphicx}
\usepackage{textcomp, comment}
\usepackage{xcolor, color}
\def\BibTeX{{\rm B\kern-.05em{\sc i\kern-.025em b}\kern-.08em
    T\kern-.1667em\lower.7ex\hbox{E}\kern-.125emX}}

\usepackage{soul}
\usepackage{float}
\usepackage{caption}
\usepackage{fontawesome,booktabs}
\usepackage{multirow}
\usepackage{pifont}
\usepackage{CJKutf8}
\usepackage{arydshln}

\begin{document}
\begin{CJK}{UTF8}{gbsn}
\newcommand{\red}[1]{\textcolor{red}{#1}}

\title{
Adapting Whisper for Code-Switching through \\ Encoding Refining and Language-Aware Decoding
}

\author{
\IEEEauthorblockN{Jiahui Zhao$^{1}$\thanks{All experiments, including all work done with the SEAME dataset, are conducted at Tianjin University and Nanyang Technological University}, Hao Shi$^{2}$, Chenrui Cui$^{1}$, Tianrui Wang$^{1}$, Hexin Liu$^{3}$, Zhaoheng Ni$^{4}$, \\ Lingxuan Ye$^{5}$, Longbiao Wang$^{1,6}$\thanks{Hao Shi and Longbiao Wang are corresponding authors.This work was supported by the National Natural Science Foundation of China under Grant 62176182 and U23B2053.}}
\IEEEauthorblockA{
{$^{1}$College of Intelligence and Computing, Tianjin University, Tianjin, China} \\
{$^{2}$Graduate School of Informatics, Kyoto University, Kyoto, Japan}\\
{$^{3}$College of Computing and Data Science, Nanyang Technological University, Singapore}
{$^4$Meta, New York, USA}\\
{$^{5}$Key Laboratory of Speech Acoustics and Content Understanding, Institute of Acoustics, Beijing, China}\\
{$^{6}$Huiyan Technology (Tianjin) Co., Ltd, Tianjin, China}}
}

\maketitle

\begin{abstract}

Code-switching (CS) automatic speech recognition (ASR) faces challenges due to the language confusion resulting from accents, auditory similarity, and seamless language switches. 
Adaptation on the pre-trained multi-lingual model has shown promising performance for CS-ASR. 
In this paper, we adapt Whisper, which is a large-scale multilingual pre-trained speech recognition model, to CS from both encoder and decoder parts. 
First, we propose an encoder refiner to enhance the encoder's capacity of intra-sentence swithching. 
Second, we propose using two sets of language-aware adapters with different language prompt embeddings to achieve language-specific decoding information in each decoder layer. 
Then, a fusion module is added to fuse the language-aware decoding. 
The experimental results using the SEAME dataset show that, compared with the baseline model, the proposed approach achieves a relative MER reduction of 4.1\% and 7.2\% on the dev\_man and dev\_sge test sets, respectively, surpassing state-of-the-art methods. 
Through experiments, we found that the proposed method significantly improves the performance on non-native language in CS speech, indicating that our approach enables Whisper to better distinguish between the two languages. 
\end{abstract}


\section{Introduction}
\label{sec:intro}
The rapid development of model pre-training significantly benefits automatic speech recognition (ASR) \cite{conneau2020unsupervised, radford2023robust, pratap2024scaling, mambazhang,liuicassp}. 
How to more effectively utilize pre-trained models to obtain better recognition performance attracts more and more attention \cite{yang2021superb,10447667}. 
There are two main methods for ASR pre-training: self-supervised learning (SSL) \cite{zhang2023google, 10542371, sunis2024,shi2024emon,zhangmamba} and weak-supervision training \cite{radford2023robust}. 
SSL uses massive unlabeled data to pre-train the model. 
Such a pre-trained model cannot directly perform ASR \cite{conneau2020unsupervised,shi2024emon2,xiang2024is}.
Thus, finetuning on labelled datasets is necessary. 
On the other hand, weak-supervision training designs rules to obtain and filter a large amount of speech-transcription data, which is used to train the neural network directly. 
In this way, the weak-supervision training-based model can directly perform ASR \cite{radford2023robust}.

Pre-training data can involve monolingual \cite{oord2018representation} and multilingual speech\cite{zhang2023google}. 
Some studies suggest that training on multilingual data often leads to higher performance \cite{conneau2020unsupervised}. 
Multilingual speech is also referred to as inter-sentence code-switching~(CS). Code-switching (CS) refers to the switching of languages within a speech signal~\cite{chua23_interspeech}. While inter-sentence code-switching data are widely used for pre-training, intra-sentence code-switching speech, which also commonly appears in multilingual communities, is rarely considered. 
Due to the seamless transition between languages and the scarcity of paired data, CS poses significant challenges to ASR systems \cite{song2022monolingual, song2022language}, even for those with multilingual capabilities. 
Adapting large pre-trained multilingual speech models to the CS ASR task is thus crucial \cite{shi2024investigation, gao24f_interspeech}.

Enabling the model to better capture language switches within the utterance better is one of the effective ways to improve CS ASR \cite{lu2020bi, tian2022lae, ma2023lae, kulkarni2023adapting}. 
This can be achieved from both the encoder and decoder. 
By updating the parameters of the additional LID-guided and CTC-guided modules, the frame-level language identity (LID) embedded in fixed Wav2Vec 2.0 \cite{baevski2020wav2vec} is combined with Connectionist Temporal Classification (CTC) to enable the joint CTC-LID framework to handle the CS \cite{tseng2021mandarin}. 
In contrast to the abundance of multilingual speech data, intra-sentence code-switching data is relatively scarce. 
As a result, adaptation with adapter modules generally exhibits comparable or even higher performance than conducting full fine-tuning on the pre-trained model for CS-ASR.
\cite{kulkarni2023adapting} leverages Massively Multilingual Speech (MMS) to handle CS by incorporating language-specific adapters after encoder layers. 
Although inserting the adapters into the encoder makes it acquire CS capabilities, the adapter's simple structure may limit the encoding capabilities, especially for capturing the boundary of language switching \cite{aditya2024attention}.

\begin{figure}[htbp]
    \noindent
    \includegraphics[width=0.48\textwidth]{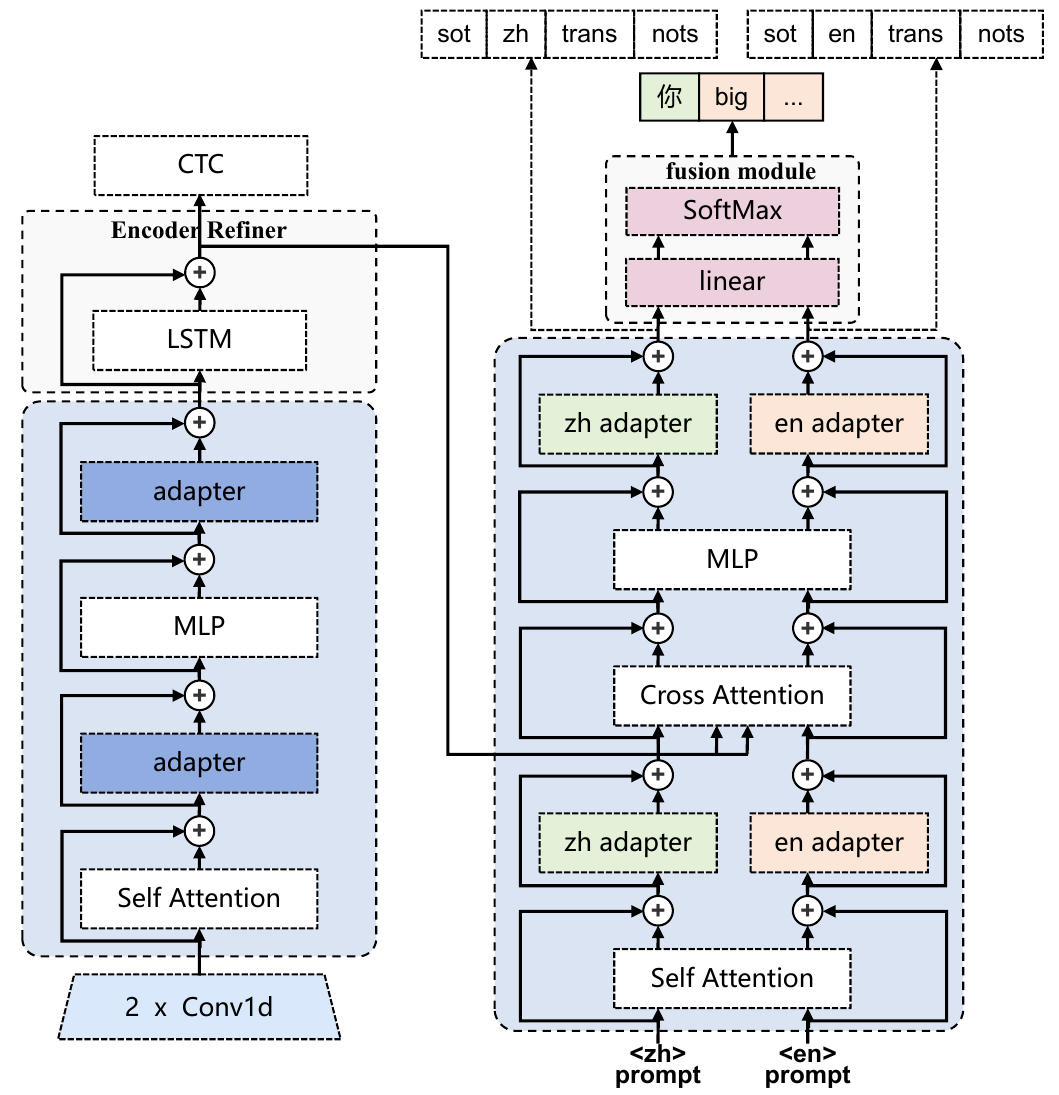}
    \vspace{-5pt}
    \caption{The encoder refiner and the language-aware adaptation of the proposed method.}
    \vspace{-15pt}

    \label{flowchart}
\end{figure}

Decoder's adaptation is often performed on Whisper \cite{radford2023robust}, which is a weakly-supervised pre-trained model for ASR and speech translation. 
In addition to using the adapter, the design of the input prompt has also shown its efficacy on Whisper performance \cite{yang2023adapting, peng2023prompting}. 
When prompts consisting of concatenated or weighted token embeddings from two languages are fed into the Whisper decoder, CS performance outperforms those with default prompts. 
However, concatenated or weighted language prompt embeddings perform poorly in zero-shot scenarios, while similar performance is observed regardless of the language prompt used during fine-tuning \cite{yang2023adapting}. Therefore, further research on prompts in the adaptation process is necessary.
Enabling Whisper to acquire better capabilities in capturing language-switching is a primary challenge in its adaptation to the CS-ASR task.

In this paper, we adapt the Whisper for CS-ASR from both the encoder and decoder modules. 
Compared with monolingual speech, CS speech is more complex, which makes the adapter needs to capture intra-sentence switching instead of only the domain mismatches. 
However, the simple adapter structure makes it hard to provide such capabilities \cite{shi2024serialized}. 
Thus, we first propose an encoding refiner, which uses a structure with stronger temporal modeling, to help the encoder obtain better CS capabilities. 
An additional CTC layer is used during training to guide the encoder refiner in learning better language-switching encoding. 
After obtaining an encoder more suitable for CS, the decoder is also designed to adapt much more efficiently. 
Different language prompts exhibit complementary characteristics in decoding \cite{peng2023prompting, yang2023adapting}. 
We thus propose language-aware decoding to investigate how to use prompts more efficiently. 
Two sets of adapters with different language prompts are inserted into each decoder layer to capture language-specific decoding information. 
A fusion module is used to fuse the two language-specific decoding embeddings after the decoder.


\section{Preliminaries}
\label{sec:preliminary}

\subsection{Adapter}
\label{ssec:adapter}
Adapter is a type of additive parameter-efficient finetuning method that involves adding domain-specific layers between neural network modules. 
In low-resource conditions, by adjusting only the parameters of the adapters, performance comparable to full finetuning can be achieved \cite{houlsby2019parameter}. 
In the field of NLP, adapters are utilized for additive parameter-efficient finetuning of large pre-trained models. 
LoRA, as a type of adapter, is typically a fully-connected network with a hidden dimension smaller than the input. 
The adaptation process of LoRA is depicted as follows: 
\begin{equation}
	\setlength{\abovedisplayskip}{2pt}
	\setlength{\belowdisplayskip}{2pt}
	\boldsymbol{e}'= \boldsymbol{e} + \text{Adapter}(\boldsymbol{e})
	\label{eq1}
\end{equation}
Here, $\boldsymbol{e}$ represents the input embedding. 
$\boldsymbol{e'}$ represents the adapted embedding.

\subsection{Low-Resource Language Adaptation}
\label{ssec:subhead}
Large-scale pre-trained multilingual ASR models have been reported to be advantageous in low-resource languages. 
Despite their excellent capabilities, these models require finetuning on target-domain data to adapt to downstream tasks \cite{liu2024exploration, wang2023adapter}. 
Finetuning large models mainly faces three challenges: being computationally and time-consuming, overfitting, and catastrophic forgetting \cite{liu2024exploration}. 
Among many finetuning strategies, adapter-based adaptation offers the dual advantage of avoiding catastrophic forgetting and reducing training time \cite{huangadapter}. 
Adapters are usually added after the attention and FFN in each Transformer encoder and decoder layer: 
\begin{equation}
	\setlength{\abovedisplayskip}{2pt}
	\setlength{\belowdisplayskip}{2pt}
    \begin{split}
     & \boldsymbol{h}^{'}  = \text{SelfAtt}(\boldsymbol{h}) + \text{Adapter}(\text{SelfAtt}(\boldsymbol{h})), \\
     & \boldsymbol{h}^{''}  = \text{MLP}(\boldsymbol{h}^{'}) + \text{Adapter}(\text{MLP}(\boldsymbol{h}^{'})), 
    \end{split}
\end{equation}
where $\boldsymbol{h}$ is the input feature or embedding from the last module of one layer. 
$\boldsymbol{h}^{'}$ and $\boldsymbol{h}^{''}$ represent the adapted embedding of the self-attention mechanism and the MLP layer.

\subsection{Whisper}
Whisper \cite{radford2023robust} is a large-scale weakly supervised pre-trained multilingual ASR model, adopting a transformer-based encoder-decoder architecture. 
Many studies on the domain adaptation of Whisper for low-resource languages ASR have already been conducted. 
Song~\textit{et al.}~\cite{song2024lora} assign a language-specific LoRA matrix for each language, enabling the model to efficiently adapt to specific languages. 
Huang~\textit{et al.}~\cite{huangadapter} integrate adapters into the Whisper model and explores their effectiveness in addressing the issue of catastrophic forgetting. 
Aditya~\textit{et al.}~\cite{aditya2024attention} employ concatenated language prompts and finetunes Whisper by incorporating adapters. 
An attention-guided Language Identification (LID) loss is introduced to enhance the decoder's ability to distinguish between the two languages. 
Peng~\textit{et al.}~\cite{peng2023prompting} and Yang~\textit{et al.}~\cite{yang2023adapting} modify Whisper prompts to adapt to CS ASR, demonstrating that $\left \langle \left | en \right |  \right \rangle$ prompt performs better on the English-dominant test set, while $\left \langle \left | zh \right |  \right \rangle$ prompt performs better on the Mandarin-dominant test set.

\section{PROPOSED Method}
\label{sec:proposed_methods}

Fig.~\ref{flowchart} illustrates the overall structure of our method, which consists of Encoding Refining and Language-aware Decoding. 
Previous works \cite{kulkarni2023adapting} show that enabling the model to better capture language switches within utterances is crucial. 
Thus, in this paper, we propose an encoder refiner to enhance the encoder's capacity  of intra-sentence swithching. 
Additionally, we propose language-aware decoding to enable Whisper's decoder to better distinguish between the two languages.

\subsection{Encoding Refining}
\label{ssec:proposed_enc}
To do the adaptation, adapters are added after each encoder layer's self-attention and the MLP to learn language switches without catastrophic forgetting. 
We propose an encoder refiner to improve the encoding capability of the Whisper encoder for ASR. 
It consists of several LSTM layers and a CTC. 
The LSTM layers have temporal modeling capability, refining the encoder's output to capture more effective encoded embeddings. 
CTC is used to train the encoder refiner to better learn language switches. 
The refining process of the encoder refiner is as follows:
\begin{equation}
	\setlength{\abovedisplayskip}{2pt}
	\setlength{\belowdisplayskip}{2pt}
    \boldsymbol{h}_{enc}^{'}  = \text{LSTM}(\boldsymbol{h}_{enc}), 
\end{equation}
where $\boldsymbol{h}_{enc}$ is the last hidden states output of the encoder. 
$\boldsymbol{h}_{enc}^{'}$ is the refined encoding. 
The training loss of the encoder refiner is as follows: 
\begin{equation}
	\setlength{\abovedisplayskip}{2pt}
	\setlength{\belowdisplayskip}{2pt}
    \mathcal{L}_{CTC} = \text{CTC}(\boldsymbol{\hat{y}}, \boldsymbol{h}_{enc}^{'}).
\end{equation}
LSTM is supervised by CTC loss, as equation, where $\boldsymbol{\hat{y}}$ represents the text tokens. 
During the finetuing, the overall training loss is defined as: 
\begin{equation}
	\setlength{\abovedisplayskip}{2pt}
	\setlength{\belowdisplayskip}{2pt}
    \mathcal{L}_{enc\_ref} = \alpha * \mathcal{L}_{att} + (1 - \alpha) * \mathcal{L}_{CTC}. 
\end{equation}
where $\mathcal{L}_{att}$ denotes the attention-based decoder loss, and $\mathcal{L}_{CTC}$ represents the CTC loss. 
The hyperparameter $\alpha$ controls the relative contribution of each loss component to the final training loss function.

\subsection{Language-aware Decoding}
\label{ssec:subhead}
To make the decoder better distinguish between the two languages, the language-aware adapters are inserted after both the self-attention and MLP of each decoder layer. 
The shared self-attention processes two language-aware embeddings: 
\begin{equation}
	\setlength{\abovedisplayskip}{2pt}
	\setlength{\belowdisplayskip}{2pt}
    \begin{split}
        &\boldsymbol{h}_{dec\_zh}  = \text{Adapter}_{zh1}(\text{SelfAtt}(\boldsymbol{p}_{zh},\boldsymbol{\hat{y}}_{1:t-1})), \\
        &\boldsymbol{h}_{dec\_en}  = \text{Adapter}_{en1}(\text{SelfAtt}(\boldsymbol{p}_{en},\boldsymbol{\hat{y}}_{1:t-1})),
    \end{split}
\end{equation}
where $\boldsymbol{p}_{zh}$, $\boldsymbol{p}_{en}$ are the language-specific prompt embeddings. 
Then, the shared frozen cross-attention processes two language-aware embeddings: 
\begin{equation}
	\setlength{\abovedisplayskip}{2pt}
	\setlength{\belowdisplayskip}{2pt}
    \begin{split}
        &\boldsymbol{h}_{dec\_zh}^{'}  = \text{CrossAtt}(\boldsymbol{h}_{dec\_zh}, \boldsymbol{h}_{enc}^{'}), \\
        &\boldsymbol{h}_{dec\_en}^{'}  = \text{CrossAtt}(\boldsymbol{h}_{dec\_en}, \boldsymbol{h}_{enc}^{'}),
    \end{split}
\end{equation}
where $\boldsymbol{h}_{enc}^{'}$ represents the output of the encoder refiner. 
Finally, the two embeedings are adapted with different adapters: 
\begin{equation}
	\setlength{\abovedisplayskip}{2pt}
	\setlength{\belowdisplayskip}{2pt}
    \begin{split}
        &\boldsymbol{h}_{dec\_zh}^{''}  = \text{Adapter}_{zh2}(\text{MLP}(\boldsymbol{h}_{dec\_zh}^{'})), \\
        &\boldsymbol{h}_{dec\_en}^{''}  = \text{Adapter}_{en2}(\text{MLP}(\boldsymbol{h}_{dec\_en}^{'})),
    \end{split}
\end{equation}
Using the language-specific adapters, which are guided by different language prompts, enables the decoder to be aware of the corresponding language during the attention-based decoding. 

At the end of the decoder, we add a lightweight fusion module. 
The fusion module uses two linear layers to estimate the weights for two language-specific embeddings: 
\begin{equation}
	\setlength{\abovedisplayskip}{2pt}
	\setlength{\belowdisplayskip}{2pt}
    \begin{split}
        &\text{weights}_{zh}^{'} = \text{Softmax}(\text{Linear}_{zh}(\boldsymbol{y}_{dec\_zh})), \\
        &\text{weights}_{en}^{'} = \text{Softmax}(\text{Linear}_{en}(\boldsymbol{y}_{dec\_en})), 
    \end{split}
\end{equation}
where $\boldsymbol{y}_{dec\_zh}$, $\boldsymbol{y}_{dec\_en}$ represents the extracted language-specific embedding of the last decoder layer. 
The $\text{weights}_{zh}^{'}$ and $\text{weights}_{en}^{'}$ are estimated for fusion: 
\begin{equation}
	\setlength{\abovedisplayskip}{2pt}
	\setlength{\belowdisplayskip}{2pt}
    \begin{split}
        & \boldsymbol{y}_{mix}^{'} = \text{weights}_{zh}^{'} * \boldsymbol{y}_{dec\_zh} + \text{weights}_{en}^{'} * \boldsymbol{y}_{dec\_en},
    \end{split}
\end{equation}
where $\boldsymbol{y}_{mix}^{'}$ represents the final fused output of decoder.The training loss for language-aware decoding is computed as follows:
\begin{equation}
	\setlength{\abovedisplayskip}{2pt}
	\setlength{\belowdisplayskip}{2pt}
    \mathcal{L}_{dec} = \mathcal{L}_{att} + \text{CE}(\boldsymbol{h}_{dec\_zh}^{''}, \boldsymbol{p}_{zh}) + \text{CE}(\boldsymbol{h}_{dec\_en}^{''}, \boldsymbol{p}_{en}).
\end{equation}
The proposed encoding refining and language-aware decoding can be used simultaneously. 
The training loss of it is as follows: 
\begin{equation}
	\setlength{\abovedisplayskip}{2pt}
	\setlength{\belowdisplayskip}{2pt}
    \mathcal{L}_{final} = \lambda * \mathcal{L}_{dec} + (1 - \lambda) * \mathcal{L}_{CTC}.
\end{equation}
$\lambda$ is a hyperparameter to balance the two losses.

\section{Experiments}
\label{sec:typestyle}

\subsection{Experimental Settings}
\label{ssec:subhead}
Our experimental data is the SEAME Mandarin-English Code-Switch dataset \cite{lyu2015mandarin}. 
Both intra- and inter-sentence CS speech exist in this dataset. 
The training set has a duration of 101.13 hours and includes 134 speakers; one of the test sets, led by Mandarin, has a duration of 7.49 hours and includes 10 speakers; one of the test sets, led by Singaporean English, has a duration of 3.93 hours and includes 10 speakers. 
Performances are measured by character error rate (CER) for Mandarin, word error rate (WER) for English, and the total mix error rate (MER) for the code-switching scenarios.

We evaluated the performance of our proposed method on Whisper-small. 
The encoder refining module was utilized with two LSTM layers. 
The hidden dimension size of LSTM was 512. 
We used Adam optimizer with learning rate $1\times 10^{-4}$. 
In the training stage, the model was trained for 8 epochs.
The evaluation model was selected from the best step based on the validation loss. 
We set $\alpha$ and $\lambda$ to 0.7 and 1.0, respectively.

\begin{table*}[htbp]
\renewcommand{\arraystretch}{1.2}
\caption{Overall performance of the proposed encoding refining and language-aware decoding for CS ASR.}
\vspace{-5pt}
\centering
\begin{tabular}{c|c|cc|cc|ccc|ccc}
\toprule[1pt]
\multirow{2}{*}{\textbf{ID}} & \multirow{2}{*}{\textbf{Experiments}} & \multicolumn{2}{c|}{\textbf{Encoder}} & \multicolumn{2}{c|}{\textbf{Decoder}} & \multicolumn{3}{c|}{\textbf{Dev\_man}} & \multicolumn{3}{c}{\textbf{Dev\_sge}} \\ \cline{3-12} 
 &  & \textbf{Adapter} & \textbf{CTC} & \textbf{Adapter} & \textbf{Prompt} & \textbf{Overall} & \textbf{ZH} & \textbf{EN} & \textbf{Overall} & \textbf{ZH} & \textbf{EN} \\ \hline
0 & \multirow{5}{*}{Baseline} & - & - & - & en/zh & 58.3 & 48.9 & 84.5 & 73.5 & 99.2 & 57.7 \\
1 &  & \checkmark &  &  & en-zh & 15.8 & 13.6 & 22.2 & 22.9 & 20.5 & 24.4 \\
2 &  &  &  & \checkmark & en-zh & 18.6 & 16.2 & 25.6 & 28.0 & 25.1 & 29.8 \\
3 &  & \checkmark &  & \checkmark & en-zh & 14.6 & 12.4 & 20.9 & 22.2 & 21.2 & 22.8 \\
4 &  & \checkmark & \checkmark & \checkmark & en-zh & 15.1 & 13.0 & 21.1 & 22.7 & 19.0 & 24.9 \\ \hline
5 & \multirow{2}{*}{\begin{tabular}[c]{@{}c@{}}Proposed:\\ Enc. Refining\end{tabular}} & \checkmark &  & \checkmark & en-zh & \red{\textbf{14.3}} & \red{\textbf{12.1}} & 20.6 & \red{\textbf{21.3}} & \red{\textbf{19.0}} & 22.7 \\
6 &  & \checkmark & \checkmark & \checkmark & en-zh & 14.4 & 12.5 & \red{\textbf{19.9}} & \red{\textbf{20.8}} & \red{\textbf{16.9}} & 23.1 \\ \hline
7 & \multirow{2}{*}{\begin{tabular}[c]{@{}c@{}}Proposed:\\ Language-aware Decoding\end{tabular}} & \checkmark &  & \checkmark & en,zh & \red{\textbf{14.1}} & \red{\textbf{12.1}} & \red{\textbf{20.0}} & \red{\textbf{20.7}} & \red{\textbf{16.6}} & 23.2 \\
8 &  & \multicolumn{2}{c|}{w/ Enc. Refining} & \checkmark & en,zh & \red{\textbf{14.0}} & 12.2 & \red{\textbf{19.2}} & \red{\textbf{20.6}} & \red{\textbf{16.6}} & 23.0 \\ 
\bottomrule[1pt]
\multicolumn{12}{r}{(Red fonts: p-value $<$ 0.05 against ID--3)} \\
\end{tabular}
\label{tab:table}
\vspace{-15pt}
\end{table*}

\begin{table}[htbp]
\centering
\caption{Comparision of  code-switching ASR on SEAME. 
``PM'' represents the large pre-trained multilingual ASR model. 
``ZT'' meas zero-shot. 
``LLM'' represents the large language model.}
\vspace{-5pt}
\renewcommand{\arraystretch}{1.2}
\begin{tabular}{c|c|c|c|c|c}
\toprule[1pt]
\multirow{2}{*}{\textbf{Methods}} &\multirow{2}{*}{\textbf{\begin{tabular}[c]{@{}c@{}}w/\\ PM\end{tabular}}} & \multirow{2}{*}{\textbf{\begin{tabular}[c]{@{}c@{}}w/\\ ZT\end{tabular}}} & \multirow{2}{*}{\textbf{\begin{tabular}[c]{@{}c@{}}w/\\ LLM\end{tabular}}} & \multicolumn{2}{c}{\textbf{MER(\%)}} \\ \cline{5-6} 
 &  &  &  & \textbf{D\_man} & \textbf{D\_sge} \\ \hline
\multicolumn{6}{c}{\textbf{Training from Scratch}} \\ \hline
Conditional CTC \cite{yang2024effective} & \ding{55} & - & \ding{55} & 18.3 & 26.7 \\
Intermediate CTC \cite{zhang2022intermediate} & \ding{55} & - & \ding{55} & 18.4 & 25.5 \\
ILME-based LM fusion \cite{peng2022internal} & \ding{55} & - & \ding{55} & 16.4 & 23.2 \\
CIF-based LSABL \cite{fan2023language} & \ding{55} & - & \ding{55} & 16.3 & 22.8 \\
token-level LID \cite{liu2023reducing} & \ding{55} & - & \ding{55} & 16.1 & 22.8 \\
CTL \cite{nga2023cyclic} & \ding{55} & - & \ding{55} & 15.6 & 22.9 \\
LAL \cite{liu2024aligning} & \ding{55} & - & \ding{55} & 16.4 & 23.3 \\
LAL + hypo. corr. \cite{liu2024aligning} & \ding{55} & - & \checkmark & 15.7 & 22.0 \\ \hline
\multicolumn{6}{c}{\textbf{Adaptation}} \\ \hline
Whisper \cite{peng2023prompting} & \checkmark & \checkmark & - & 38.2 & 65.0 \\
concated Whisper \cite{peng2023prompting} & \checkmark & \checkmark & - & 32.7 & 47.6 \\
joint CTC-LID \cite{tseng2021mandarin} & \checkmark & \ding{55} & - & 18.8 & 28.5 \\
AG Whisper \cite{aditya2024attention} & \checkmark & \ding{55} & - & 14.2 & 20.8 \\ \hline
Our Propose & \checkmark & \ding{55} & - & 14.0 & 20.6 \\ 
\bottomrule[1pt]
\end{tabular}
\label{tab:table1}
\vspace{-15pt}
\end{table}

\subsection{Experimental Results and Analysis}
\label{ssec:subhead}
Table~\ref{tab:table} shows the overall performance of encoding refining and language-aware decoding for CS ASR. 
The baselines include the pre-trained Whisper-small (ID--0) and adapter-based adaptation methods (ID--1 to ID--4). 
The decoding language prompts of the pre-trained Whisper-small (ID--0) for the dev\_man and dev\_sge test sets were $\left \langle \left | en \right |  \right \rangle$ and $\left \langle \left | zh \right |  \right \rangle$, respectively. 
Its poor performance suggests that handling CS directly using the pre-trained Whisper is challenging. 
With adapter-based adaptation, the CS performance significantly improved. 
The language prompt of all the adapter-based baselines was the concatenation of $\left \langle \left | en \right |  \right \rangle$ and $\left \langle \left | zh \right |  \right \rangle$. 
Only encoder (ID--1) or decoder (ID--2) adaptation and both encoder and decoder adaptation (ID--3) were evaluated. 
Experimental results show that the efficacy of adaptation on the encoder was better than that on the decoder. 
Moreover, the adaptation of both encoder and decoder further improved performance. 
We also used CTC to improve the encoder misalignment problem further (ID--4). 
However, this degraded the model performance. 

\subsubsection{Performance of Encoding Refining}
\label{sssec:subsubhead}
The result of encoding refining is shown in Table \ref{tab:table} (ID--5). 
Its language prompt was also the concatenation of $\left \langle \left | en \right |  \right \rangle$ and $\left \langle \left | zh \right |  \right \rangle$. 
Compared with the best result in baselines (ID--3), the encoder refiner (ID--5 and ID--6) significantly improved the performance. 
Most interesting is that the encoder refining improved the performance of non-native speaker regions, especially training with CTC (ID--6): 
the relative improvement was about 5\% for EN in Dev\_man subset; while the relative improvement was about 20\% in ZH in Dev\_sge subset. 
ID--4 and ID--6 all trained with CTC. 
The encoder refiner allows the model to learn more information from CTC using a more complex neural network structure instead of only the LoRA adapter.

\subsubsection{Performance of Language-aware Decoding}
\label{sssec:subsubhead}
The proposed language-aware decoding uses two sets of adapters for processing different language information. 
Experimental results (ID--7) showed that the proposed language-aware decoding also improved the performance of non-native speaker regions: the relative improvement was about 4\% for EN in Dev\_man subset; while the relative improvement was about 22\% in ZH in Dev\_sge subset. 
Using the proposed encoder refiner and language-aware decoding simultaneously further improved performance (ID--8). 
Overall, the proposed method (ID--8) had about 4.1\% and 7.2\% relatively improvement on the Dev\_man, and Dev\_sge sets over the baseline model (ID--3), respectively.

\subsubsection{Comparison with other Outstanding Methods}
\label{sssec:subsubhead}
Table \ref{tab:table1} compares different CS ASR models on the SEAME dataset in the literature. 
Those well-performing models that were trained from scratch and adapted based on the pre-trained models were all used for comparison. 
Models trained from scratch are not zero-shot on CS. 
Thus, they had good CS ability, especially on the Dev\_man set. 
Among these methods, aligning different language parts was effective, especially after adding a speech model (``LAL'' and ``LAL + hypo. corr.''). 
The pre-trained models are often zero-shot. 
Although some content can be recognized when evaluating CS directly on these pre-trained models, the performance is relatively poor. 
Using CS data to adapt the models significantly improved the pre-trained model performance. 
AG Whisper \cite{aditya2024attention} represents the previous state-of-the-art model before our proposed method.

\section{CONCLUSIONS}
\label{sec:majhead}

In this work, we propose an encoder refiner for the Whisper encoder and perform language-aware adaptation on the Whisper decoder. The encoder refiner can adopt various structures, and we demonstrated its effectiveness through experiments with LSTM and CTC architectures. Experimental results show that the encoder refiner achieves a 2.1\% and 6.3\% relative MER reduction on dev\_man and dev\_sge test sets compared to the baseline model by enhancing the encoder's capability to encode CS speech. Additionally, in language-aware decoding, the two sets of adapters within each decoder layer extract language-specific information under different language prompts. Experimental results show that language-aware decoding achieves a 3.4\% and 6.8\% relative MER reduction on dev\_man and dev\_sge test sets compared to the baseline model. This demonstrates that the two points we proposed for adapting Whisper are effective, and the final model achieves a 4.1\% and 7.2\% relative MER reduction on dev\_man and dev\_sge test sets compared to the baseline model.

\end{CJK}

\clearpage

\begin{thebibliography}{00}

\bibitem{conneau2020unsupervised} A. Conneau, A. Baevski, R. Collobert, A. Mohamed, and M. Auli, ``Unsupervised cross-lingual representation learning for speech recognition,'' in Proc. Interspeech, 2021, pp. 2426--2430.

\bibitem{radford2023robust} A. Radford, J.-W. Kim, T. Xu, G. Brockman, C. McLeavey, and I. Sutskever, ``Robust speech recognition via large-scale weak supervision,'' in Proc. ICML, 2023, pp. 28492--28518.

\bibitem{pratap2024scaling} V. Pratap, A. Tjandra, B. Shi, P. Tomasello, A. Babu, S. Kundu, A. Elkahky, Z. Ni, A. Vyas, M. Fazel-Zarandi, A. Baevski, Y. Adi, X. Zhang, W.-N. Hsu, A. Conneau, and M. Auli, ``Scaling speech technology to 1,000+ languages,'' Journal of Machine Learning Research, vol. 25, no. 97, pp. 1--52, 2024.

\bibitem{mambazhang} X. Zhang, Q. Zhang, H. Liu, T. Xiao, X. Qian, B. Ahmed, E. Ambikairajah, H. Li, and J. Epps, ``Mamba in Speech: Towards an Alternative to Self-Attention,'' arXiv preprint arXiv:2405.12609, 2024. 

\bibitem{liuicassp} H. Liu, L. P. Garcia, X. Zhang, A. W. H. Khong and S. Khudanpur, "Enhancing Code-Switching Speech Recognition With Interactive Language Biases," in Proc. ICASSP, 2024, pp. 10886-10890. 


\bibitem{yang2021superb} S.-w. Yang, P.-H. Chi, Y.-S. Chuang, et al, ``SUPERB: Speech processing universal performance benchmark,'' in Proc. Interspeech, 2021, pp. 1194-1198.

\bibitem{10447667} H. Sun, S. Zhao, X. Wang, W. Zeng, Y. Chen and Y. Qin, "Fine-Grained Disentangled Representation Learning For Multimodal Emotion Recognition," in Proc. ICASSP, 2024, pp. 11051-11055. 

\bibitem{zhang2023google} Y. Zhang, W. Han, J. Qin, et al, ``Google USM: Scaling automatic speech recognition beyond 100 languages,'' arXiv preprint arXiv:2303.01037, 2023.

\bibitem{10542371} H. Shi, M. Mimura, and T. Kawahara, ``Waveform-domain speech enhancement using spectrogram encoding for robust speech recognition,'' IEEE/ACM Trans. on Audio, Speech, and Language Processing, vol. 32, pp. 3049--3060, 2024.

\bibitem{zhangmamba} X. Zhang, J. Ma, M. Shahin, B. Ahmed, and J. Epps, “Rethinking
mamba in speech processing by self-supervised models,” arXiv preprint
arXiv:2409.07273, 2024.

\bibitem{sunis2024} H. Sun, S. Zhao, X. Kong, X. Wang, H. Wang, J. Zhou,
and Y. Qin, “Iterative prototype refinement for ambiguous speech emotion recognition,” in Interspeech, 2024, pp. 3200–3204.

\bibitem{shi2024emon} X. Shi, J. He, X. Li, and T. Toda, ``On the effectiveness of asr representations in real-world noisy speech emotion recognition,'' arXiv preprint arXiv:2311.07093.

\bibitem{xiang2024is} S. Dang, T. Matsumoto, Y. Takeuchi, T. Tsuboi, Y. Tanaka, D. Nakatsubo, S. Maesawa, R. Saito, M. Katsuno, and H. Kudo, ``Developing vocal system impaired patient-aimed voice quality assessment approach using ASR representation-included multiple features'' in Proc. Interspeech, 2024, pp.2445--2449. 

\bibitem{shi2024emon2} X. Shi, X. Li, and T. Toda, ``Multimodal Fusion of Music Theory-Inspired and Self-Supervised Representations for Improved Emotion Recognition,'' in Proc. Interspeech, 2024, pp.2024--2350.

\bibitem{xiang2024apsipa} S. Dang, T. Matsumoto, Y. Takeuchi, H. Kudo, T. Tsuboi, Y. Tanaka, and M. Katsuno, "Using Self-learning Representations for Objective Assessment of Patient Voice in Dysphonia," in Proc. APSIPA ASC, 2022, pp. 359-363.

\bibitem{oord2018representation} A. v. d. Oord, Y. Li, and O. Vinyals, ``Representation learning with contrastive predictive coding,'' arXiv preprint arXiv:1807.03748, 2018.

\bibitem{chua23_interspeech} V. Y. H. Chua, H. Liu, L. P. Garcia, F. T. Woon, J. Wong, X. Zhang, S. Khudanpur, A. W. H. Khong, J. Dauwels, and S. J. Styles, ``MERLIon CCS Challenge: A English-Mandarin code-switching child-directed speech corpus for language identification and diarization,'' in Proc. Interspeech, 2023, pp. 4109--4113.

\bibitem{song2022monolingual} T. Song, Q. Xu, H. Lu, L. Wang, H. Shi, Y. Lin, Y. Yang, and J. Dang, ``Monolingual recognizers fusion for code-switching speech recognition,'' arXiv preprint arXiv:2211.01046, 2022.

\bibitem{song2022language} T. Song, Q. Xu, M. Ge, L. Wang, H. Shi, Y. Lv, Y. Lin, and J. Dang, ``Language-specific characteristic assistance for code-switching speech recognition,'' in Proc. Interspeech, 2022, pp. 3924--3928.

\bibitem{shi2024investigation} H. Shi and T. Kawahara, ``Exploration of adapter for noise robust automatic speech recognition,'' arXiv preprint arXiv:2402.18275, 2024.

\bibitem{gao24f_interspeech} Y. Gao, H. Shi, C. Chu, and T. Kawahara, ``Speech emotion recognition with multi-level acoustic and semantic information extraction and interaction,'' in Proc. Interspeech, 2024, pp. 1060--1064.

\bibitem{kulkarni2023adapting} A. Kulkarni, A. Kulkarni, M. Couceiro, and H. Aldarmaki, ``Adapting the adapters for code-switching in multilingual asr,'' arXiv preprint arXiv:2310.07423, 2023.

\bibitem{lu2020bi} Y. Lu, M. Huang, H. Li, J. Guo, and Y. Qian, ``Bi-encoder transformer network for mandarin-english code-switching speech recognition using mixture of experts,'' in Proc. Interspeech, 2020, pp. 4766--4770.

\bibitem{tian2022lae} J. Tian, J. Yu, C. Zhang, C. Weng, Y. Zou, and D. Yu, ``LAE: Language-aware encoder for monolingual and multilingual asr,'' in Proc. Interspeech, 2022, pp. 3178--3182.

\bibitem{ma2023lae} G. Ma, W. Wang, Y. Li, Y. Yang, B. Du, and H. Fu, ``LAE-ST-MOE: Boosted language-aware encoder using speech translation auxiliary task for E2E code-switching ASR,'' in Proc. ASRU, 2023, pp. 1--8.

\bibitem{baevski2020wav2vec} A. Baevski, Y. Zhou, A. Mohamed, and M. Auli, ``wav2vec 2.0: A framework for self-supervised learning of speech representations,'' Advances in neural information processing systems, vol. 33, pp. 12449--12460, 2020.

\bibitem{tseng2021mandarin} L.-H. Tseng, Y.-K. Fu, H.-J. Chang, and H.-y. Lee, ``Mandarin-english code-switching speech recognition with self-supervised speech representation models,'' arXiv preprint arXiv:2110.03504, 2021.

\bibitem{aditya2024attention} B. Aditya, M. Rohmatillah, L.-H. Tai, and J.-T. Chien, ``Attention-guided adaptation for code-switching speech recognition,'' in Proc. ICASSP, 2024, pp. 10256--10260.

\bibitem{shi2024serialized} H. Shi, Y. Gao, Z. Ni, and T. Kawahara, ``Serialized speech information guidance with overlapped encoding separation for multi-speaker automatic speech recognition,'' arXiv preprint arXiv:2409.00815, 2024.

\bibitem{peng2023prompting} P. Peng, B. Yan, S. Watanabe, and D. Harwath, ``Prompting the hidden talent of web-scale speech models for zero-shot task generalization,'' in Proc. Interspeech, 2023, pp. 396--400.

\bibitem{yang2023adapting} Y. Yang, Y. Peng, X. Zhong, H. Huang, and E.-S. Chng, ``Adapting openai's whisper for speech recognition on code-switch mandarin-english seame and asru2019 datasets,'' arXiv preprint arXiv:2311.17382, 2023.

\bibitem{houlsby2019parameter} N. Houlsby, A. Giurgiu, S. Jastrzebski, B. Morrone, Q. de Laroussilhe, A. Gesmundo, M. Attariyan, and S. Gelly, ``Parameter-efficient transfer learning for nlp,'' in Proc. ICML, 2019, pp. 2790--2799.

\bibitem{liu2024exploration} Y. Liu, X. Yang, and D. Qu, ``Exploration of whisper fine-tuning strategies for low-resource asr,'' EURASIP Journal on Audio, Speech, and Music Processing, vol. 2024, no. 1, pp. 29, 2024.

\bibitem{wang2023adapter} T. Wang, X. Chen, Z. Chen, S. Yu, and W. Zhu, ``An adapter based multi-label pre-training for speech separation and enhancement,'' in Proc. ICASSP, 2023, pp. 1--5.

\bibitem{huangadapter} Z. Huang, H. Xing, and M. Liu, ``Adapter Integration: Mitigating catastrophic forgetting in multi-language and multi-accent whisper asr model fine-tuning,'' .

\bibitem{song2024lora} Z. Song, J. Zhuo, Y. Yang, Z. Ma, S. Zhang, and X. Chen, ``LoRA-Whisper: Parameter-efficient and extensible multilingual asr,'' arXiv preprint arXiv:2406.06619, 2024.

\bibitem{lyu2015mandarin} D.-C. Lyu, T.-P. Tan, E.-S. Chng, and H. Li, ``Mandarin--english code-switching speech corpus in south-east asia: Seame,'' Language Resources and Evaluation, vol. 49, pp. 581--600, 2015.

\bibitem{peng2022internal} Y. Peng, Y. Liu, J. Zhang, H. Xu, Y. He, H. Huang, and E. S. Chng, ``Internal language model estimation based language model fusion for cross-domain code-switching speech recognition,'' arXiv preprint arXiv:2207.04176, 2022.

\bibitem{fan2023language} Z. Fan, L. Dong, C. Shen, Z. Liang, J. Zhang, L. Lu, and Z. Ma, ``Language-specific acoustic boundary learning for mandarin-english code-switching speech recognition,'' arXiv preprint arXiv:2306.05279, 2023.

\bibitem{liu2023reducing} H. Liu, H. Xu, L. P. Garcia, A. WH Khong, Y. He, and S. Khudanpur, ``Reducing language confusion for code-switching speech recognition with token-level language diarization,'' in Proc. ICASSP, 2023, pp. 1--5.

\bibitem{nga2023cyclic} C. H. Nga, D.-Q. Vu, H. H. Luong, C.-L. Huang, and J.-C. Wang, ``Cyclic transfer learning for mandarin-english code-switching speech recognition,'' IEEE Signal Processing Letters, 2023.

\bibitem{liu2024aligning} H. Liu, X. Zhang, L. P. Garcia, A. WH Khong, E. S. Chng, and S. Watanabe, ``Aligning speech to languages to enhance code-switching speech recognition,'' arXiv preprint arXiv:2403.05887, 2024.

\bibitem{yang2024effective} T.-T. Yang, H.-W. Wang, Y.-C. Wang, C.-H. Lin, and B. Chen, ``An effective mixture-of-experts approach for code-switching speech recognition leveraging encoder disentanglement,'' in Proc. ICASSP, 2024, pp. 11226--11230.

\bibitem{zhang2022intermediate} J. Zhang, Y. Peng, H. Xu, Y. He, E. S. Chng, and H. Huang, ``Intermediate-layer output regularization for attention-based speech recognition with shared decoder,'' arXiv preprint arXiv:2207.04177, 2022.




\end{thebibliography}
\end{document}